\documentclass[lettersize,journal]{IEEEtran}
\usepackage{amsmath,amsfonts}
\usepackage{algorithmic}
\usepackage{algorithm}
\usepackage{array}
\usepackage[caption=false,font=footnotesize,labelfont=rm,textfont=rm]{subfig}
\usepackage{textcomp}
\usepackage{stfloats}
\usepackage{url}
\usepackage{verbatim}
\usepackage{graphicx}
\usepackage{cite}
\usepackage{tabularx}
\usepackage{multirow}
\usepackage{threeparttable}
\usepackage{amsmath}
\usepackage{amssymb}
\usepackage{amsthm}
\usepackage{booktabs}
\begin{document}

\title{Global Priors Guided Modulation Network for Joint Super-Resolution and Inverse Tone-Mapping}
\author{Gang He, Shaoyi Long, Li Xu, Chang Wu, Jinjia Zhou, Ming Sun, Xing Wen, Yurong Dai
\thanks{Gang He, Shaoyi Long, Li Xu, Chang Wu are with the State Key Laboratory of Integrated Services Networks, School
of Telecommunications Engineering, Xidian University, Xi'an 710071, Shaanxi, P. R. China (e-mail: ghe@xidian.edu.cn; sylong@stu.xidian.edu.cn; cherylxu@stu.xidian.edu.cn;  Jackwu0630@outlook.com). (Corresponding author: Shaoyi Long.)}
\thanks{Jinjia Zhou is with Hosei University, Koganei, Tokyo 184-8584, Japan (e-mail: jinjia.zhou.35@hosei.ac.jp).}
\thanks{Ming Sun, Xing Wen, Yurong Dai are with Kuaishou Technology (e-mail: sunming03@kuaishou.com; td.wenxing@gmail.com; yuwing@gmail.com)
}
}

\markboth{}%
{He \MakeLowercase{\textit{et al.}}: Global Priors Guided Modulation Network for Joint Super-Resolution and Inverse Tone-Mapping}


\maketitle

\begin{abstract}
Joint super-resolution and inverse tone-mapping (SR-ITM) aims to enhance the visual quality of videos that have quality deficiencies in resolution and dynamic range. This problem arises when using 4K high dynamic range (HDR) TVs to watch a low-resolution standard dynamic range (LR SDR) video. Previous methods that rely on learning local information typically cannot do well in preserving color conformity and long-range structural similarity, resulting in unnatural color transition and texture artifacts. In order to tackle these challenges, we propose a global priors guided modulation network (GPGMNet) for joint SR-ITM. In particular, we design a global priors extraction module (GPEM) to extract color conformity prior and structural similarity prior that are beneficial for ITM and SR tasks, respectively. To further exploit the global priors and preserve spatial information, we devise multiple global priors guided spatial-wise modulation blocks (GSMBs) with a few parameters for intermediate feature modulation, in which the modulation parameters are generated by the shared global priors and the spatial features map from the spatial pyramid convolution block (SPCB). With these elaborate designs, the GPGMNet can achieve higher visual quality with lower computational complexity. Extensive experiments demonstrate that our proposed GPGMNet is superior to the state-of-the-art methods. Specifically, our proposed model exceeds the state-of-the-art by 0.64 dB in PSNR, with 69$\%$ fewer parameters and 3.1$\times$ speedup. The code will be released soon.
\end{abstract}


\section{Introduction}
\IEEEPARstart{C}{ompared} with low-resolution (LR) and standard dynamic range (SDR) videos, ultra-high definition (UHD) and high dynamic range (HDR) videos can more realistically represent natural scenes with higher resolution, deeper bit depth and richer colors from a wider color gamut (WCG). They can be viewed on 4K/8K HDR TVs or other devices that support UHD resolution and HDR formats including HDR10, HDR10+, Dolby Vision, and HLG. The popularity of UHD HDR display devices are increasing in our daily life, but vast majority of video sources are still in LR SDR format. In order to provide the best user watching experiences on UHD HDR displays, it is necessary to convert LR SDR sources to their UHD HDR version.

This task, the joint SR and ITM processing, is of great practical value and is gradually receiving attention in the research community of image reconstruction. Software solutions can restore older LR SDR videos to their higher quality versions to enhance people's visual experience and adapt to UHD HDR display devices. In addition, the bandwidth can be reduced by directly reconstructing the transmitted LR SDR video to HR HDR on the local device or cloud server using a software solution. Therefore, it is significant to design an appropriate algorithm for Joint SR-ITM.

\begin{figure}[!t]
\centering
\includegraphics[width=\linewidth]{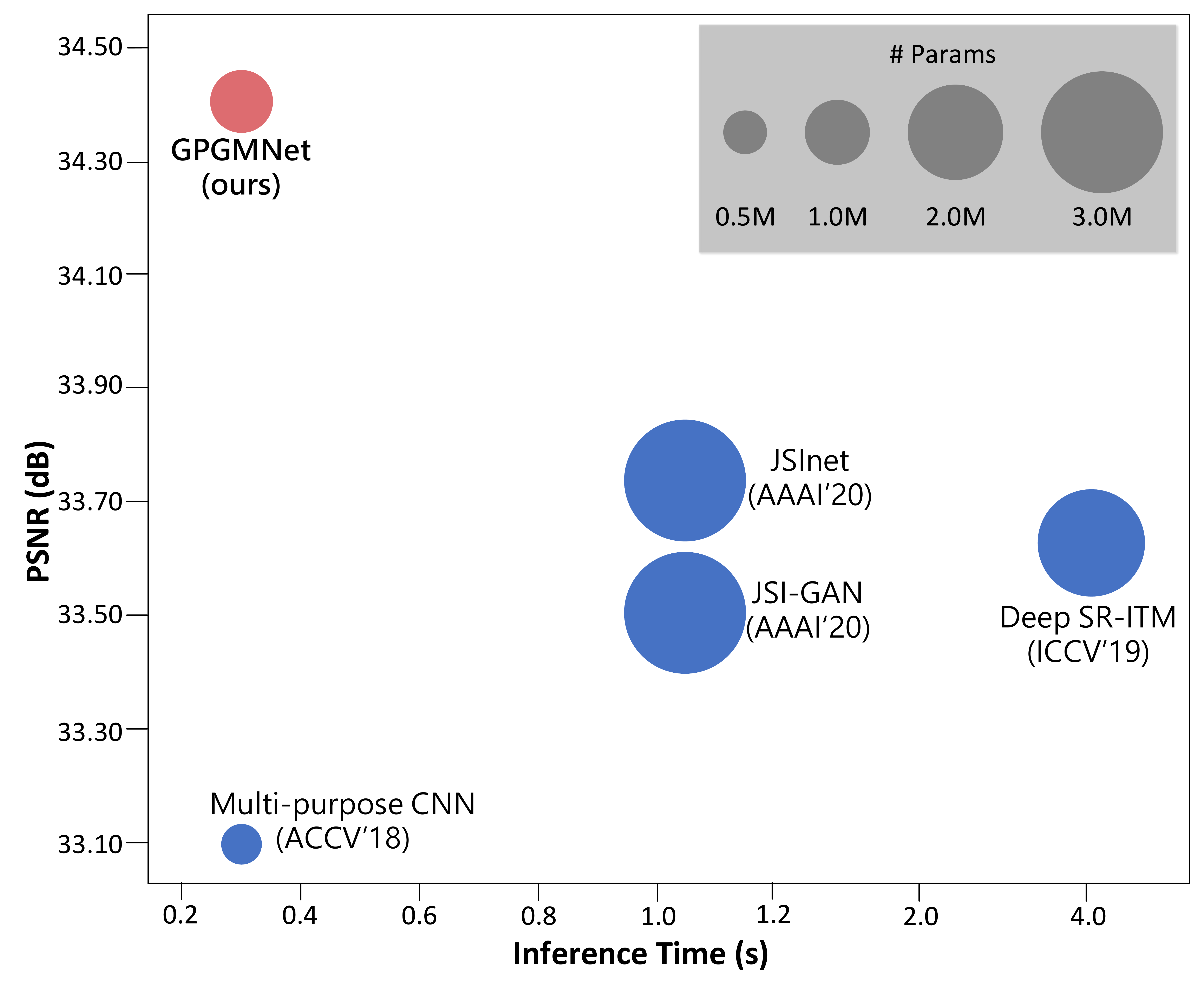}
\caption{Performance and speed comparison. Our proposed GPGMNet outperforms existing state-of-the-arts with much lower computational cost on the 4K HDR dataset \cite{kim2019deep}. The inference time for 4$\times$ upsampling is computed with an NVIDIA TITAN Xp GPU on an LR SDR size of 540$\times$960.}
\label{fig1}
\end{figure}

\begin{figure}[!t]
\centering
\includegraphics[width=\linewidth]{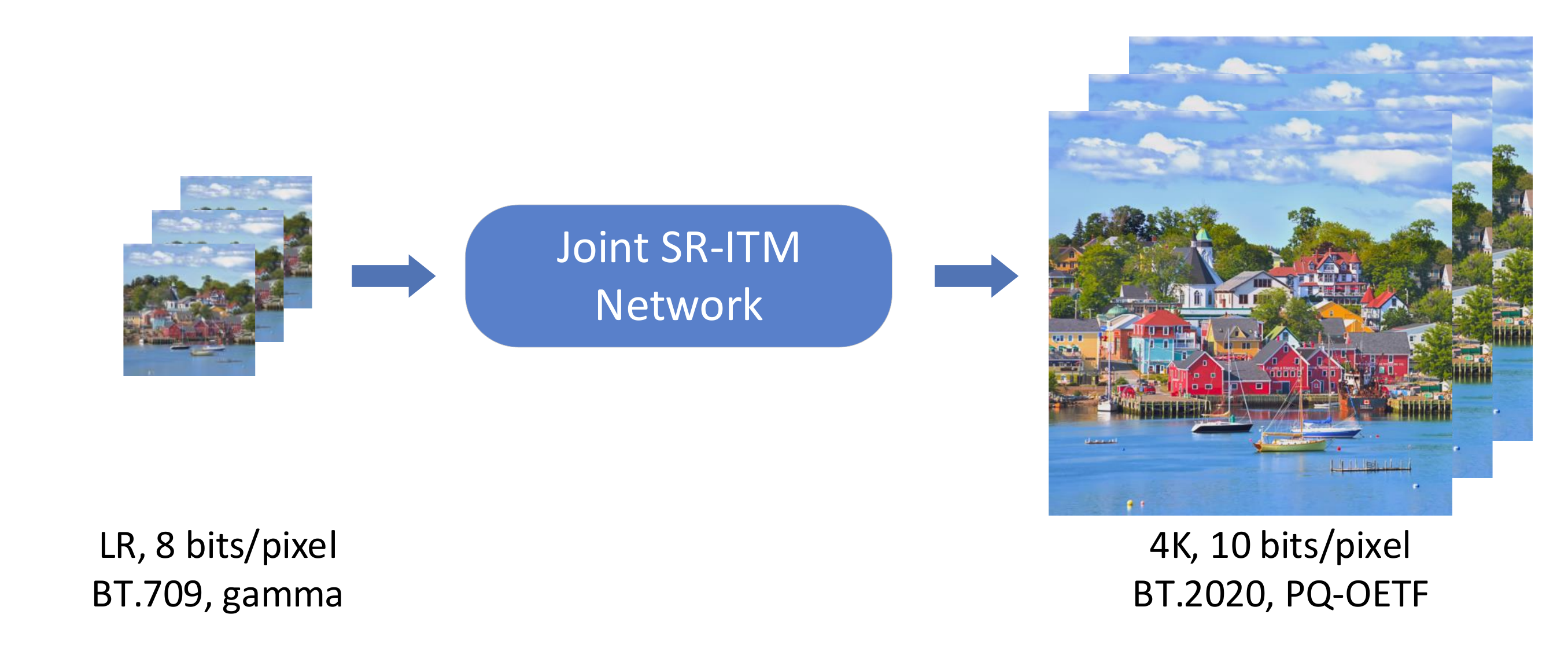}
\caption{Illustrates the mapping of LR-SDR frames to 4K-HDR frames in resolution, bit depth, color container and transfer function, respectively. Joint SR-ITM network aims to map LR-SDR video to the corresponding 4K-HDR version in an end-to-end manner.}
\label{fig2}
\end{figure}

Joint SR-ITM is a complex and tricky problem. For SR task, it needs to restore high frequency information and texture details. For ITM task, it needs to enhance local contrast, extend bit depth and generate highlight details. Figure \ref{fig2} illustrates the four types of mappings that need to be implemented in the Joint SR-ITM network. Directly connecting SR and ITM methods cannot effectively exploit the common information and the same processing steps between the two tasks. Additionally, the independent two stage methods may accumulate errors from the previous prediction \cite{kim2018multi, 9622212, 9620041}. Recently, Deep SR-ITM \cite{kim2019deep} decomposed the input as base and detail by using guided filter, and then jointly restore high frequency details and local contrast while increasing the spatial resolution and extending dynamic range. JSI-GAN \cite{kim2020jsi} designed three task-specific sub-networks with GAN framework for better visual quality and artifact removal. Although kim et al. \cite{kim2019deep, kim2020jsi} achieved state-of-the-art results, their methods could generate many texture and transaction artifacts. In addition, their methods require a large computational cost when aiming to recover 4K images. The earlier method Multi-purpose CNN \cite{kim2018multi} has fewer parameters and faster inference time, but its performance is lower than those of JSI-GAN \cite{kim2020jsi} and Deep SR-ITM \cite{kim2019deep} in the comparison of qualitative and quantitative results. 

Overall, it is still challenging to design an end-to-end deep network with high performance and efficiency for reconstructing 4K HDR videos. To tackle this problem, we propose a global priors modulated framework, denoted as GPGMNet, to cooperatively modulate color information and texture details. First, the Joint SR-ITM task requires some ITM-specific global operations, such as expanding the color gamut from BT.709 to BT.2020 \cite{series2012parameter} and mapping the opto-electronic transfer function (OETF) from gamma to PQ-OETF \cite{standard2014high} for HDR10 standard. Additionally, SR-specific reconstructions also require global information, including structural similarity prior and global image statistics. Therefore, we design the global priors extraction module (GPEM) to adequately extract the global priors through color conformity prior and structural similarity prior extraction branches. Preserving spatial information is also important for the Joint SR-ITM task, which usually requires adaptive processing at different spatial locations of an image, such as enhancing local contrast and avoiding homogeneous feature modulation. To effectively leverage global priors and preserving spatial information, we ingeniously design the joint modulated residual module (JMRM), which mainly contains a spatial pyramid convolution block (SPCB) and a global priors guided spatial-wise modulation block (GSMB). The SPCB extracts the multi-scale spatial features and broadcasts them to the GSMB, which generates spatial-wise modulation parameters by incorporating shared global priors and aggregated spatial information, followed by the scaling and shifting operations. Benefiting from a compressed global priors vector and an aggregated spatial feature map, our proposed GPGMNet achieves significant feature modulation with a small computational cost. Finally, the generated 4K HDR reconstruction are obtained by adding the residual of the predicted image to a up-sampled image. With these elaborate designs, our models can obtain more substantial quantitative and qualitative results while saving computational costs. Several methods and the comparison of our method are shown in Figure \ref{fig1}.

Our contributions are summarized as:
\begin{itemize}
\item We present a global priors extraction module with two task-specific branches for joint SR-ITM problem. The GPEM can continuously refine the global priors through ITM-specific and SR-specific branches.
\item We propose the joint modulated residual modules, which mainly consists of a GSMB and a SPCB. The GSMB with a few parameters generates spatial-wise modulation parameters by incorporating shared global priors and aggregated spatial information for feature modulation.
\item With these novel elaborate designs, GPGMNet surpass existing state-of-the-arts by 0.64 dB in PSNR. In addition, our model achieves faster inference speed and less number of parameters.
\end{itemize}

\section{Related Work}
\subsection{Joint SR-ITM}
Early work utilized the multi-purpose CNN \cite{kim2018multi} approach to perform SR, ITM and SR-ITM tasks simultaneously in a single network with three output separately. However, this approach is not tailor-made for the task feature of SR and ITM. Kim et al \cite{kim2019deep} proposed a more sophisticated end-to-end based approach Deep SR-ITM to decompose the input into detail and base information for spatially-variant operations. Based on the idea of decomposing the input, Kim et al. proposed the JSInet \cite{kim2020jsi}, which designs three task-specific sub-networks, including detail restoration subnet, local contrast enhancement subnet and image reconstruction subnet. Furthermore, kim et al. \cite{kim2020jsi} introduced a GAN-based approach, denoted as JSI-GAN, to further improve the perceptual quality of the HR HDR frames generated by JSInet. Although JSI-GAN achieves better visual quality than Deep SR-ITM in artifact removal and texture detail recovery, it still suffers from edge artifacts, inaccurate color mapping and unnatural transitions. A recent approach similar to Joint SR-ITM was proposed by Tan et al. \cite{9633177}. They designed the Deep SR-HDRI network and demonstrated the effectiveness of the joint network over the independent two-stage network with extensive experiments. Since the purpose of the Deep SR-HDRI network is to generate HDR images in the linear domain using information from multiple frames, it is not suitable for the Joint SR- ITM task. In order to handle these challenges in the task, we propose a global priors guided modulation framework to reconstruct a HR HDR frame by jointly learning and modulating color information and texture details.

\subsection{Super-Resolution}
Starting from Dong et al.'s SRCNN \cite{dong2015image}, deep learning approaches have made great progress in image super-resolution \cite{ledig2016photo, lim2017enhanced, zhang2018residual, kim2016accurate, dong2016accelerating, lai2017deep, zhang2018image, 9400479, 9233990, 9044197, 9614658, 9705493}. Soon, Kim et al. proposed DRCN and VDSR with deeper networks on residual learning. Further, Lin et al. \cite{ledig2016photo} introduced residual block into SSResNet. EDSR \cite{lim2017enhanced} greatly improved the performance of SR by a simple and meaningful change - removing the Batch Normalization (BN) layers in SSResNet. Based on EDSR, Zhang et al. \cite{zhang2018residual} introduced densely connected blocks to the residual blocks to form the residual dense blocks (RDN). After that, they introduced channel attention mechanism and non-local blocks into the residual network to form the residual channel attention network (RCAN) \cite{zhang2018image} and residual non-local attention network (RNAN) \cite{zhang2019residual}, respectively. Furthermore, Hu et al. \cite{8708220} introduce channel-wise and spatial feature modulation blocks to dynamically modulate the feature maps in global and local manners. With the considerable performance of the residual block, it is also used in many advanced SR frameworks \cite{wang2019edvr, chan2021basicvsr, chan2021basicvsr++}. However, these methods, which do not consider the ITM component, are not suitable for directly applying in the Joint SR-ITM task. More recently, Liu et al. \cite{9663185} propose a multi-scale feature fusion method for preserving structural information of images. Different with them, we propose more lightweight spatial pyramidal convolution blocks to extract multi-scale spatial features and then fuse it with structural similarity priors information for further modulation.

\begin{figure*}[t]
  \centering
  \includegraphics[width=\linewidth]{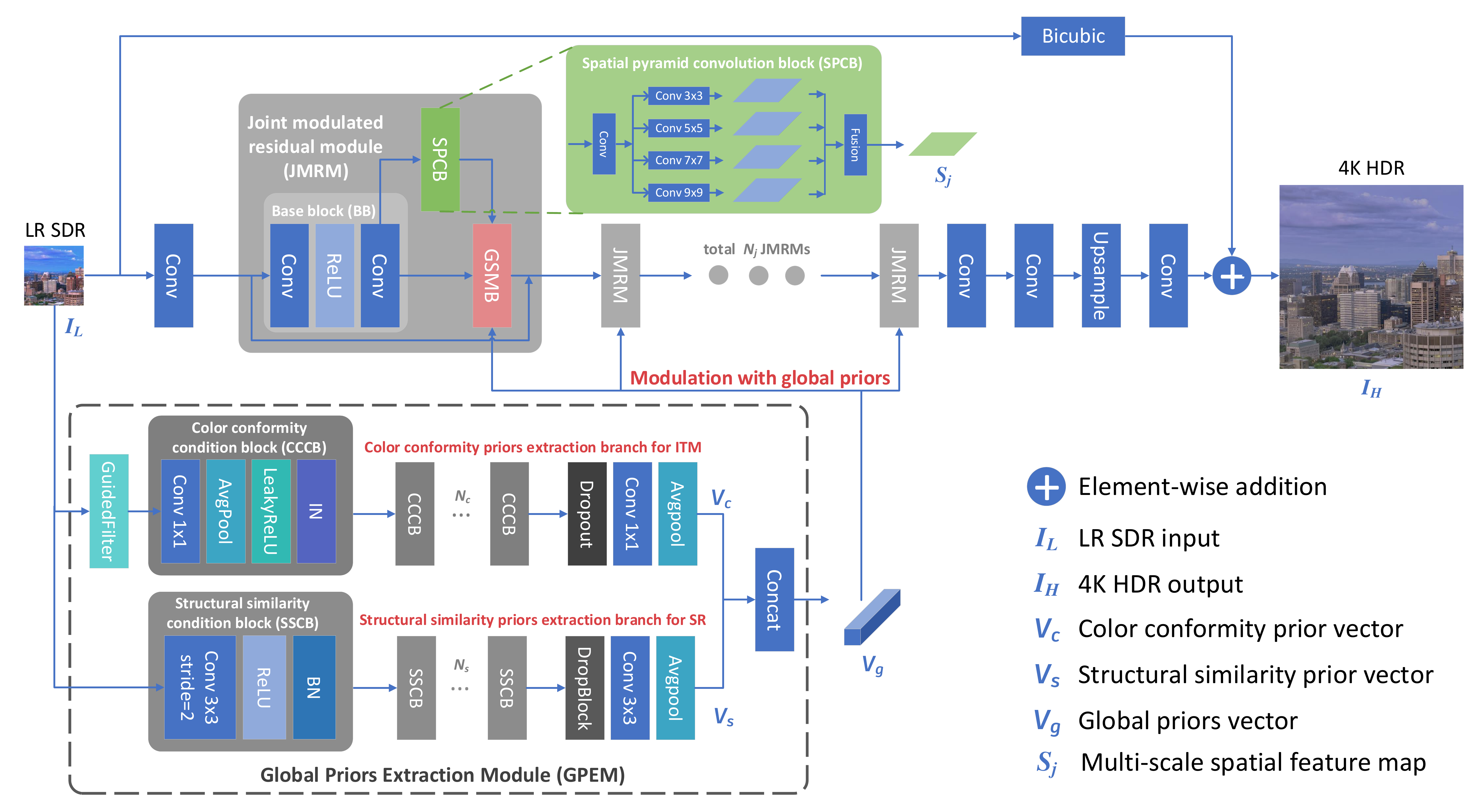}
  \vspace{-0.25in}
  \caption{Network architecture of the proposed GPGMNet. It predicts a 4K HDR frame from a single frame of low-resolution standard dynamic range (LR SDR) video. All JMRMs share a global priors vector from the GPEM, which generates an intermediate global priors from the input and transmits the priors to the GSMB (Figure \ref{fig4}) for further modulation of the feature maps.}
  \label{fig3}
\end{figure*}

\subsection{Inverse Tone-Mapping}
In this study, the purpose of ITM is to convert SDR video to its HDR version. CNN-based single-exposure ITM methods \cite{EKDMU17, marnerides2018expandnet, 7572895, liu2020single} were often used in the past to predict linear HDR radiograms from LDR images. These methods are inappropriate for directly recovering HDR videos with PQ-OETF. Recently, some ITM methods \cite{8611162, 8759906, 9311239, chen2021new} have been developed in order to recover HDR videos from the corresponding SDR videos. Chen et al. \cite{chen2021new} analyzed the SDR video to HDR video formation pipeline and proposed the HDRTVNet, a cascaded approach containing three deep networks for SDRTV-to-HDRTV problem. HDRTVNet implemented HDR video recovery with a three-stage training strategy. First, AGCM is used to implement color mapping. Second, an EDSR-like local enhancement network implements local enhancement. Finally, the highlight generation network uses the GAN framework to enhance the visual quality of highlight regions. In particular, AGCM achieves a comparable performance to Ada-3DLUT \cite{zeng2020learning} by extracting global priors and modulating these priors with GFM \cite{he2020conditional}. However, this strategy of directly modulating the global priors without considering spatial information may lead to artifacts and texture distortions in the joint SR-ITM problem. Although cascaded EDSR and AGCM can handle the joint SR-ITM problem, this approach is inefficient and cannot exploit the common information between the SR and ITM tasks.

\section{APPROACH}
In this work, the aim of joint SR-ITM is to estimate its counterpart 4K HDR video from a single frame of LR SDR video. In order to produce high quality results, it is important to restore fine texture, enhance local contrast and adjust chroma information while maintaining correct color mapping, smooth color transitions and consistent structural similarity. Recent studies that rely on learning local features focus on recovering different features with different sub-networks without considering global priors, which result in inappropriate holistic color, transition artifacts and texture problem. It is worth thinking about how to extract global priors and how to use these priors. Furthermore, maintaining spatial information is crucial for joint SR-ITM, since different spatial locations of an image require different processing. For handle joint SR-ITM problem, we propose an end-to-end network based on the CNN framework, called GPGMNet, which fully considers the global priors and performs spatial-wise modulation with these priors.

\subsection{Network Overview}
An overview of GPGMNet is depicted in Figure \ref{fig3}. It is a simple, powerful and effective network. The network mainly consists of several joint modulated residual modules (JMRMs), a global priors extraction module (GPEM), an upsampling module, convolution layers and skip connection. The details of the GPEM and a JMRM components are described in Sec. III-B and III-C. The upsampling operation is performed after the last convolution of the network to improve the spatial resolution and thus save the computation. Finally, the generated 4K HDR frame $I_{H}$ is obtained by adding the residual of the predicted image $I_{L}$ to the upsampled image. Skip connection is used to enhance the prediction accuracy. Our architecture can be formulated as:
\begin{equation}
\begin{split}
I_{H} = & CL \circ Up \circ CB \circ \\ 
& JMRM^{N_{j}}\{GPEM(I_L), CF(I_L)\} \\
& + Bic(I_L)
\label{deqn_ex1a}
\end{split}
\end{equation}
where $N_j$ denotes the number of the JMRM, $CF$ denote the first convolution, $CB$ denotes the convolution block, $CL$ denotes the last convolution, and $Bic$ denotes the bicubic interpolation algorithm. The GPEM generates global priors and broadcasts it to JMRMs. All JMRMs shared a global priors vector.

\subsection{Global Priors Extraction Module}
For the joint SR-ITM task, the key to maintaining color conformity and structural similarity is to extract global priors that benefit both ITM and SR tasks simultaneously. Different images have different holistic colors, specific structures and global image statistics. To fully extract priors, we propose a GPEM with color conformity prior (CCP) and structural similarity prior (SSP) extraction branches. CSRNet \cite{he2020conditional} extracts the global priors using $N_k \times N_k(N_k>1)$ filters for image retouching. Furthermore, AGCM \cite{chen2021new} uses $1 \times 1$  filters to focus on extracting global color priors for the SDRTV-to-HDRTV task. Thus, inspired by these methods, we introduce the CCP extraction branch for ITM. Unlike AGCM, we perform a guided filtering \cite{he2012guided} of the low-resolution frame before feeding it to the convolution layers. This operation allows the CCP extraction branch to more intensively extract the CCP and low-frequency information. For SR and HDR image reconstruction task, SFTGAN \cite{wang2018recovering} and HDRUNet \cite{chen2021hdrunet} use a condition network to extract spatial feature or location-specific priors from the segmentation probability maps and input image, respectively. However, these methods do not focus on compressing and extracting the global priors and require more computational cost. Therefore, we propose the SSP extraction branch for SR. The SSP extraction branch extracts the SSP and plausible high-frequency information by continuously compressing and refining the overall structural features into a one-dimensional vector. Finally, we concatenate the output of these two branches and broadcast it to JMRMs. The proposed global priors extraction module is shown in Figure \ref{fig3}.

\vspace{2mm}
\noindent\textbf{Color Conformity Prior Extraction Branch. } The branch consists of a guided filtering, several color condition blocks (CCCBs), dropout \cite{srivastava2014dropout}, convolution layers with $1 \times 1$ filters and global average pooling. Our CCP extraction branch takes a LR SDR image $I_L$ as input and outputs a CCP vector $V_c$. The branch can be written as follows:
\begin{equation}
V_c = GAP \circ Conv_{1\times1} \circ Dropout \circ CCCB^{N_c} \circ GF(I_L)
\end{equation}
where GAP is global average pooling, $Conv_{k\times k}$ denotes a convolution layer with $k\times k$ kernels, $N_c$ denotes the number of color condition blocks, and $GF$ denotes guided filtering. Specifically, a CCCB can be described as:
\begin{equation}
  CCCB(\cdot) = IN \circ LReLU \circ avgpool \circ Conv_{1\times1}(\cdot)  
\end{equation}
where $IN$ is instance normalization \cite{huang2017arbitrary}, LReLU is Leaky ReLU activation \cite{maas2013rectifier}, $(\cdot)$ denotes the input of CCCB. Note that the CCCB and color condition block in AGCM have the same structure.

\vspace{2mm}
\noindent\textbf{Structural Similarity Prior Extraction Branch.} The branch includes several structural similarity condition blocks (SSCBs), convolution layers with $3\times3$ filters, DropBlock \cite{ghiasi2018dropblock}, and global average pooling. It takes a $I_L$ as input and generates a SSP vector $V_{s}$. The SSP extraction branch is denoted by:
\begin{equation}
   V_s = GAP \circ Conv_{3\times3} \circ DropBlock \circ SSCB^{N_s}(I_s) 
\end{equation}
where $N_s$ denotes the number of SSCB. A SSCB can described as:
\begin{equation}
   SSCB(\cdot) = BN \circ ReLU \circ Conv_{3\times3}(\cdot) 
\end{equation}
where $BN$ is batch normalization \cite{ioffe2015batch}. To avoid overfitting, we introduce DropBlock with $3\times3$ filters. Experiments show that DropBlock is more appropriate than Dropout in SSP extraction branch.

Finally, we concatenate the CCP vector $V_c$ and the SSP vector $V_{s}$ to obtain a global priors vector $V_g$. This can be indicated as:
\begin{equation}
   V_g = [V_c \quad V_{s}] 
\end{equation}

\begin{figure} [t!]
	\centering
	\subfloat[Spatial Feature Transform Block \cite{wang2018recovering, chen2021hdrunet}\label{1a}]{
		\includegraphics[width=0.95\linewidth]{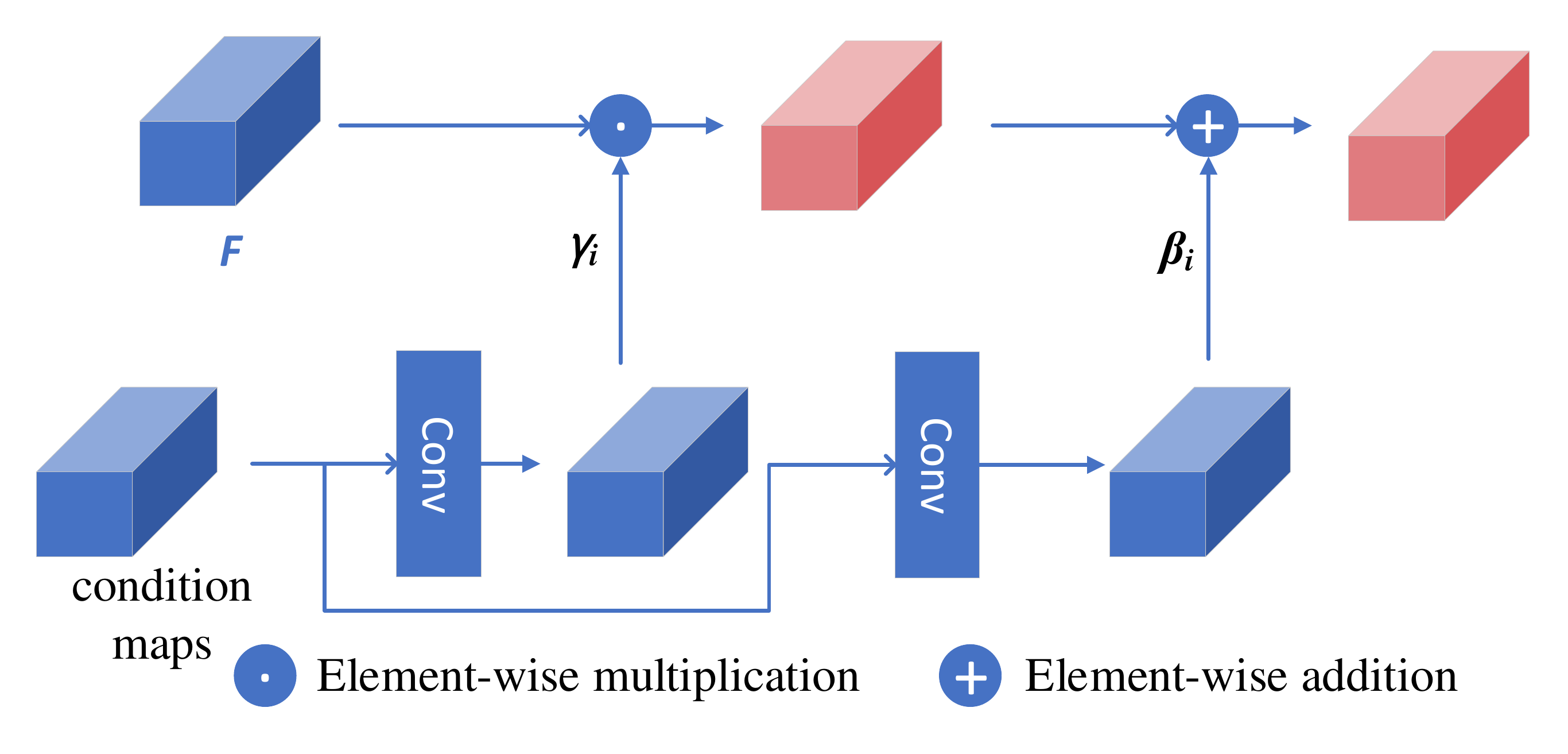}}
	\\
	\subfloat[Global Feature Modulation Block \cite{he2020conditional, chen2021new}\label{1a}]{
		\includegraphics[width=0.95\linewidth]{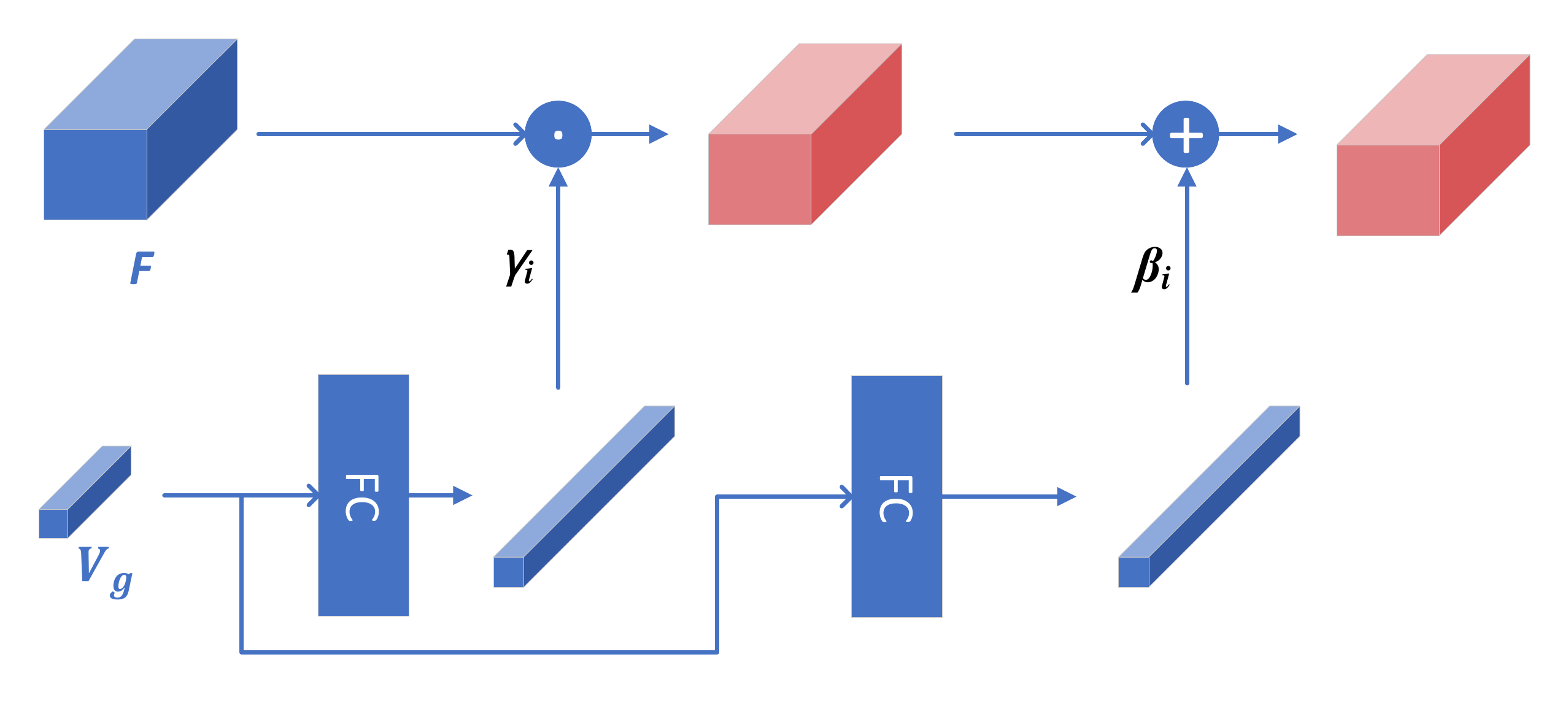}}
	\\
	\subfloat[Global priors guided Spatial-wise Modulation Block (Ours)\label{1b}]{
		\includegraphics[width=0.95\linewidth]{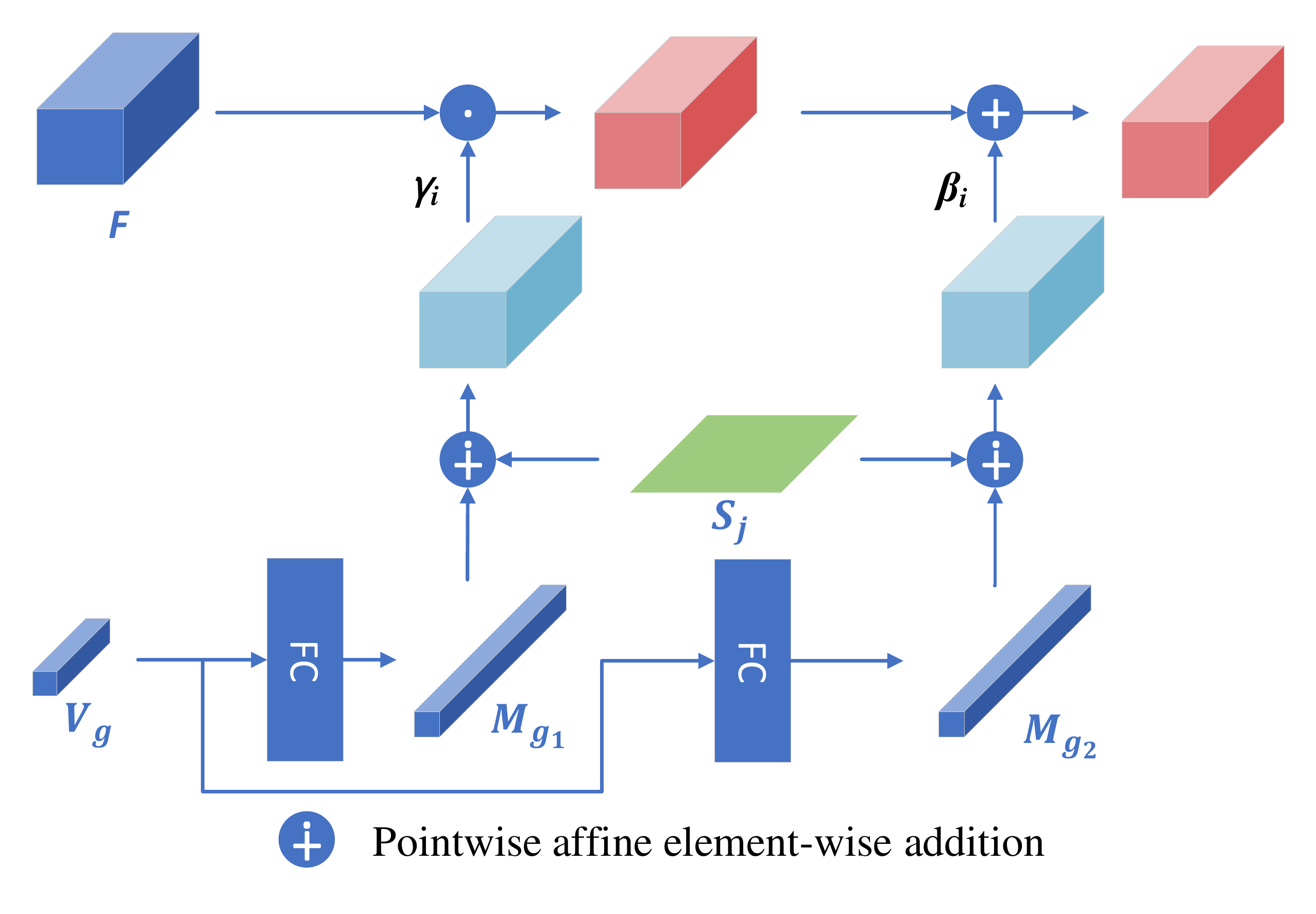} }
	\caption{Modulation block structure comparison. (a) Spatial feature transform \cite{wang2018recovering, chen2021hdrunet}, which performs feature-wise modulation. (b) Global feature modulation block \cite{he2020conditional, chen2021new}, which modulates the feature map directly using global priors. (c) Our proposed global priors guided spatial-wise modulation block (GSMB), which performs spatial-wise modulation on feature maps after fusing spatial location information.}
	\label{fig4} 
\end{figure}

\subsection{Joint Modulated Reconstruction Module}
Both super-resolution and inverse tone mapping belong to the field of low-level vision, and these two tasks are not independent of each other. They both need to reconstruct high-frequency information and low-frequency information. In addition, preserving spatial information is crucial for low-level tasks. Therefore, we propose the joint modulated residual modules (JMRMs) for the joint SR-ITM problem, in order to restore chroma information, enhance local contrast and reconstruct texture details efficiently. A JMRM consists of a base block (BB), a spatial pyramid convolution block (SPCB), and a global priors guided spatial-wise modulation block (GSMB). First, the base block restores the basic high and low frequency information , extracts local features and transmits the spatial features. Then, the SPCB further extracts the spatial information and broadcasts it to the GSMB. Finally, the GSMB collaboratively modulate the color information and texture details by scaling and shifting operations.

\vspace{2mm}
\noindent\textbf{Base Block.} The base block comprise of two convolution layers with $3\times3$ kernels and a ReLU activation function, which can be expressed as:
\begin{equation}
   BB(\cdot) = Conv_{3\times3} \circ ReLU \circ Conv_{3\times3}(\cdot) 
\end{equation}

\vspace{2mm}
\noindent\textbf{Spatial Pyramid Convolution Block.} Inspired by spatial pyramid pooling (SPP) \cite{he2015spatial} and spatial attention (SA) module \cite{woo2018cbam}, we design a spatial pyramid convolution block (SPCB) for each JMRMs. Similar to SPP, it extracts multi-scale features through convolutional layers with kernels of different sizes. Unlike SA, SPCB retains a small number of learnable parameters while reducing the number of channels to 1, which enables SPCB to perform adaptive processing in each JMRMs. Compared with a single convolution layer, the SPCB can extract spatial information more sufficiently by diverse perceptual fields. It takes feature maps $F \in \mathbb{R}^{C\times H\times W}$ as input and outputs a spatial feature map $S_{j} \in \mathbb{R}^{1\times H\times W}$, as shown in Figure \ref{fig3}. The SPCB can be denoted as:   
\begin{equation}
\begin{split}
    S_{j} &= Fusion \circ [Conv_{3\times3}\, Conv_{5\times5}\,Conv_{7\times7}\,Conv_{9\times9}] \\
    & \circ Conv(\textbf{F})
\end{split}
\end{equation}
where $F$ denotes the feature maps and $Fusion$ includes a concatenation operation and a convolution layer. The different features are fused into a spatial feature $S_{j}$ with 1 channel by the $Fusion$ operation. The SPCB extracts the spatial features efficiently with a little number of parameters. 

\vspace{2mm}
\noindent\textbf{Global Priors Guided Spatial-wise Modulation Block.}  Different areas in one image have different holistic tone, color, local contrast, texture and spatial features. For joint SR-ITM problem, it is necessary to perform global mapping and spatial-wise modulation. Therefore, we revisit spatial feature transform (SFT) \cite{wang2018recovering, chen2021hdrunet} and global feature modulation (GFM) \cite{he2020conditional, chen2021new}. The SFT generates a pair $(\gamma, \beta) \in \mathbb{R}^{C\times H\times W}$ by a condition maps, and then performs spatial modulations through scaling and shifting, respectively. The SFT can be presented as:
\begin{equation}
    SFT(\textbf{F}|cond) = \gamma \odot \textbf{F} \oplus \beta
\end{equation}
where $\odot$ is referred to element-wise multiplication and $\oplus$ is element-wise addition. $cond$ is condition maps, which is generated by  condition network \cite{chen2021hdrunet}. 

The GFM generate a pair $(\gamma, \beta) \in \mathbb{R}^{C\times 1\times 1}$ by a global priors vector, and then conducts global feature modulations through scaling and shifting, respectively. The GFM can be given as:
\begin{equation}
    GFM(\textbf{F}|V_g) = \gamma \odot \textbf{F} \oplus \beta
\end{equation}
where $V_g$ is global priors vector, which is generated by  condition network \cite{chen2021new}. 

The SFT performs feature-wise modulation ignoring the global mapping. The GFM applies global modulation and cannot maintain spatial location information. For joint SR-ITM problem, directly modulating the global priors is inefficient
and leads to homogeneous spatial feature modulation. To efficiently leverage global priors and preserve spatial information, we propose the \textbf{G}lobal priors guided \textbf{S}patial-wise \textbf{M}odulation Block (GSMB), which takes feature maps, a spatial feature map and a global priors vector as input and outputs modulated features. First, the GSMB learns a mapping function $M$ that generates a pair of global priors guided maps $(M_{g_1}, M_{g_2}) \in \mathbb{R}^{C\times 1\times 1}$ based on a global priors vector $V_g$ . Second, the guided maps generate a learnable modulation parameter pair $(\gamma, \beta) \in \mathbb{R}^{C\times H\times W}$ by applying a pointwise affine element-addition to each spatial pixel of the spatial feature map $S_{j} \in \mathbb{R}^{1\times H\times W}$. Finally, the parameter pair is used to modulate each intermediate feature maps $F \in \mathbb{R}^{C\times H\times W}$ through scaling and shifting operations. The overall guided modulation mechanism can be summarized as:
\begin{subequations}\label{eq:2}
\begin{align}
(M_{g_1}, M_{g_2})&=M(V_g) \label{eq:2A}\\
(\gamma, \beta) &=(M_{g_1}, M_{g_2}) \dotplus S_{j} \label{eq:2B}\\
GSMB(\textbf{F}|V_g,S_{j}) &= \gamma \odot \textbf{F} \oplus \beta \label{eq:2c}
\end{align}
\end{subequations}
where $\dotplus$ denotes a pointwise affine element-addition. During pointwise affine element-addition, the guided maps and the spatial feature map are broadcasted (copied) accordingly: guided maps are broadcasted along the spatial dimension, and the sptail feature map is broadcasted along the channel dimension. 

Figure \ref{fig4} shows the different structures and modulations of SFT, GFM and our proposed GSMB. In comparison to SFT, which is focused on spatial features, and GFM, which is based on global prior, we implement global priors guided spatial-wise modulation, which simultaneously leverage global priors and spatial location information. In particular, we use two independent fully-connected layers as the mapping function $M$ so that it can generate two $M_{g_i}$ from a global priors vector of arbitrary dimensions in a content-aware manner. The $M_{g_i}$ share a $S_{j}$ and GSMBs share a $V_g$ extracted by GPEM for efficiency. We keep few parameters of the mapping function inside each GSMB to adapt a specific $M_{g_i}$ to further guide a specific $S_{j}$, thus providing different parameters pairs $(\gamma, \beta)$ to modulate the intermediate feature maps. 

\begin{table*}
  \caption{Quantitative comparison with previous method on 4K HDR dataset \cite{kim2019deep}. All results are calculated on YUV-channel. The inference time (Inf. Time) for 2$\times$ and 4$\times$ upsampling is computed on a LR SDR of size 1080$\times$1920 and 540$\times$960, respectively.}
  \label{tab1}
  \begin{tabular}{ccccccccc}
    \toprule
    Method &Scale &PSNR (dB)$\uparrow$ &SSIM$\uparrow$ &mPSNR (dB)$\uparrow$ &MS-SSIM$\uparrow$ &HDR-VDP2$\uparrow$ &Inf. Time (s)$\downarrow$ &Params (M)$\downarrow$\\
    \midrule
    EDSR+Huo et al. &$\times$2 & 29.76 &0.8934 &31.81 & 0.9764 &58.95  &- &-  \\
    EDSR+Eilertsen et al. &$\times$2 & 25.80 &0.7586 &28.22 &0.9635 &53.51 &- &- \\
    Multi-purpose CNN (ACCV'18) &$\times$2 &34.11 &0.9671 &36.38 &0.9817 &{60.91} &\textbf{0.49} &\textbf{0.25} \\
    Deep SR-ITM (ICCV'19) &$\times$2 & 35.58 &0.9746 &37.80 &0.9839 &\textbf{61.39} &5.02 &2.50 \\
    JSInet (AAAI'20) &$\times$2 & {35.99} &{0.9768} &{38.20} &{0.9843} &60.58 &1.72 &1.45 \\
    JSI-GAN (AAAI'20) &$\times$2 & 35.73 &0.9763 &37.96 & 0.9841 &60.80 &1.72 &1.45 \\
    \textbf{GPGMNet (Ours)} &$\times$2 &\textbf{36.29} &\textbf{0.9784} &\textbf{38.44} &\textbf{0.9866} &60.34 &{0.78} &{0.77} \\
    \midrule
    EDSR+Huo et al. &$\times$4 & 28.90 &0.8934 &31.81 & 0.9693 &55.59 &- &-  \\
    EDSR+Eilertsen et al. &$\times$4 & 26.54 &0.7822 &28.75 &0.9631 &53.88 &- &- \\
    Multi-purpose CNN (ACCV'18) &$\times$4 & 33.10 &0.9499 &35.26 &0.9758 &56.41 &\textbf{0.34} &\textbf{0.28} \\
    Deep SR-ITM (ICCV'19) & $\times$4 & 33.61 &0.9561 &35.73 &0.9748 &56.07 &4.06 &2.64 \\
    JSInet (AAAI'20) & $\times$4 &{33.74} &{0.9598} &{35.93} &{0.9759} &{56.45} &1.05 &3.03 \\
    JSI-GAN (AAAI'20) & $\times$4 & 33.50 &0.9572 &34.82 & 0.9743 &56.41 &1.05 &3.03 \\
    \textbf{GPGMNet (Ours)} &$\times$4 &\textbf{34.38} &\textbf{0.9635} &\textbf{36.76} &\textbf{0.9791} &\textbf{57.23}  &\textbf{0.34} &{0.92} \\
    \bottomrule
  \end{tabular}
\end{table*}

Overall, a JMRM can be formulated as (\ref{eq10}). Benefiting from JMRMs and global priors extraction module, GPGMNet has achieved outstanding performance. 
\begin{equation}
\label{eq10}
   JMRM(\textbf{F},V) = GSMB\{BB(\textbf{F})|V, SPCB \circ BB(\textbf{F})\} + \textbf{F} 
\end{equation}

\begin{table}
  \centering
  \caption{Quantitative comparison with JSInet and JSI-GAN on the test set of HDRTV1K dataset \cite{chen2021new}. Our proposed GPGMNet also achieves state-of-the-art results in PSNR and SSIM.}
  \label{tab2}
  \setlength{\tabcolsep}{5mm}{
  \begin{tabular}{ccc}
    \toprule
    Method &PSNR (dB) &SSIM\\
    \midrule
    JSInet (AAAI'20) &28.36 &0.8686  \\
    \midrule
    JSI-GAN (AAAI'20) &28.38 &0.8557  \\
    \midrule
    \textbf{GPGMNet (Ours)} &\textbf{29.08} &\textbf{0.8857}  \\
    \bottomrule
    \end{tabular}}
\end{table}

\begin{table}[]
\centering
\caption{Quantitative comparison with the deep cascade method. We directly connect the state-of-the-art SR method, RLFN, and the state-of-the-art ITM method for HDR video (HDR10), AGCM (ICCV 2021) \cite{chen2021new}. We train and test the cascade method and our proposed GPGMNet on the 4K HDR dataset \cite{kim2019deep}.}
\label{tab3}
\setlength{\tabcolsep}{5mm}{
  \begin{tabular}{ccc}
    \toprule
    Method &PSNR (dB) & Param (K) \\ \midrule
    \begin{tabular}[c]{@{}c@{}}RLFN+AGCM\\ (SR + ITM)\end{tabular} & 33.87 & 939   \\ \midrule
    \begin{tabular}[c]{@{}c@{}}AGCM+RLFN\\ (ITM + SR)\end{tabular} & 33.88 & 939   \\ \midrule
    \begin{tabular}[c]{@{}c@{}}\textbf{GPGMNet}\\ \textbf{(Ours)}\end{tabular} & \textbf{34.38} & \textbf{920}   \\
    \bottomrule
    \end{tabular}}
\end{table}


\begin{figure*}[h]
  \centering
  \includegraphics[width=\linewidth]{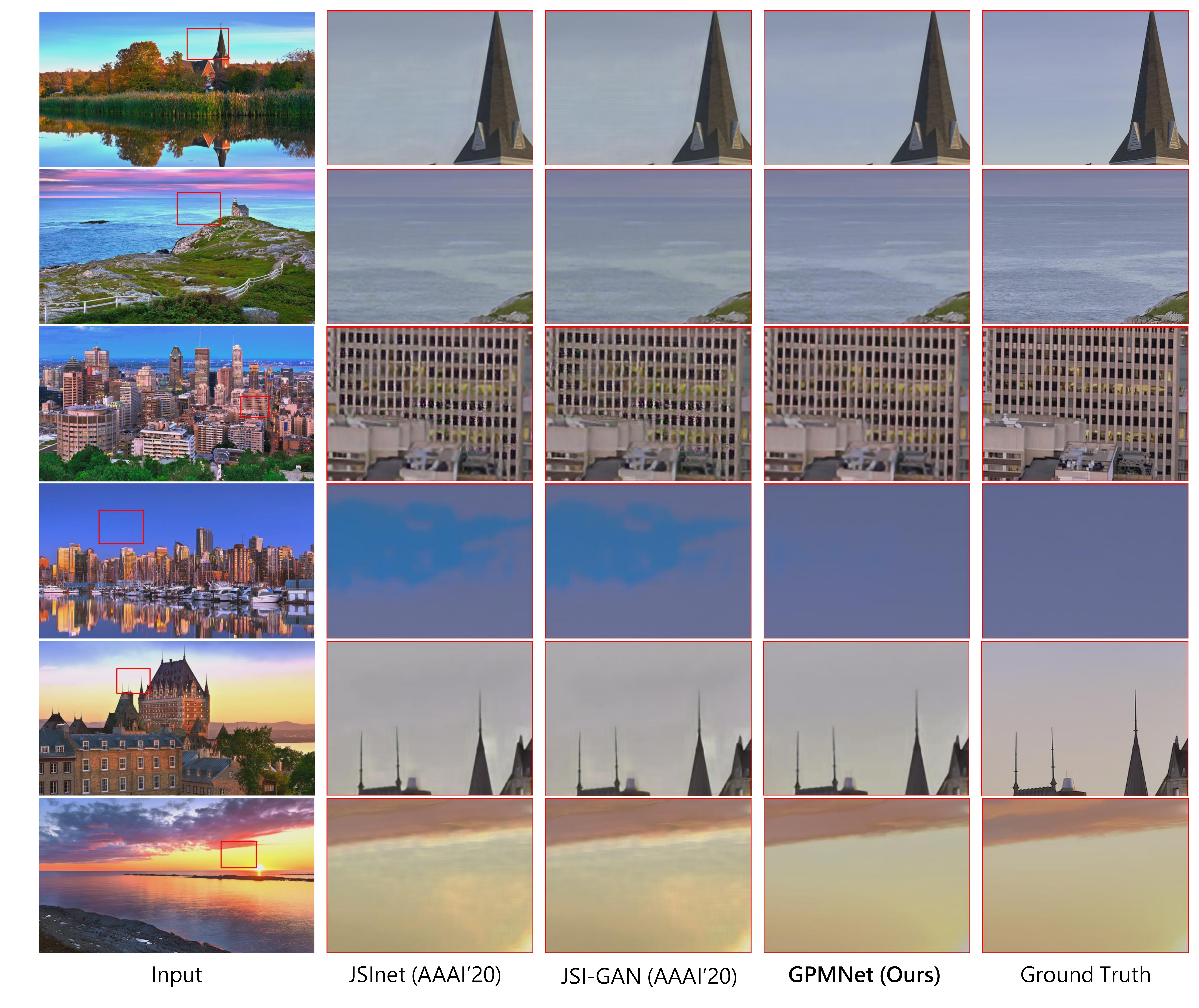}
  \vspace{-0.25in}
  \caption{Visual qualitative comparison. Part of the areas are zoomed in and framed with a red box to facilitate comparison. }
  \vspace{-0.1in}
  \label{fig5}
\end{figure*}

\section{EXPERIMENTS}
\subsection{Experiments Detail}
\noindent\textbf{Datasets.} We utilize the 4K HDR dataset collected and produced by Kim et al. \cite{kim2019deep, kim2020jsi}. They gathered 4K HDR videos (BT.2020 with PQ OETF) and their corresponding 4K SDR videos from YouTube, and then obtained SDR video pairs by downsampling with bicubic filtering. For training, the ground truth patches of 160x160 size were cropped from 10 bit YUV frames (BT.2020 with PQ OEFT) and the LR SDR patches of 40x40 or 80x80 size were extracted from the corresponding 8 bit YUV frames (BT.709). For testing, we used 28 different scenes following \cite{kim2019deep, kim2020jsi}.

\vspace{1mm}
\noindent\textbf{Training.} For the proposed GPGMNet, the global priors extraction module contains 5 CCCBs and 3 SSCBs, and the number of the joint modulated residual modules is 5 with 64 channels. We set the dimensions of both the CCP vector and the SSP vector to 6. For scales $\times2$ and $\times4$, we employ 1 pixelshuffle and 2 pixelshuffles respectively in the upsampling module. We used the L2 loss and Adam \cite{kingma2014adam} optimizer for training following kim et al. \cite{kim2019deep, kim2020jsi}. The initialized learning rate is 1e4, the batch size is 32, and the total number of iterations is 290K.

\subsection{Comparisons}
We compare our proposed GPGMNet with three previous Joint SR-ITM methods, the Multi-purpose CNN, the Deep SR-ITM, the JSInet, and the JSI-GAN. Note that JSInet is trained solely on the L2 loss without the GAN framework, and JSI-GAN is trained on the Adversarial loss and the detail GAN loss with the GAN framework. The GPGMNet and these methods are trained on the same dataset. The GPGMNet is also compared with a cascade of an SR methods, EDSR \cite{lim2017enhanced}, and two ITM methods, Huo et al. \cite{huo2014physiological} and Eilertsen et al. \cite{eilertsen2017hdr}, following Kim et al. \cite{kim2020jsi}. Neither of these two ITM methods is based on deep learning, and they cannot be trained on 4K HDR datasets. . For comparison with CNN-based cascade methods in an end-to-end manner, we designed an deep cascade method for SR-ITM task. All data-driven methods are re-trained on the same dataset.

\vspace{2mm}
\noindent\textbf{Evaluation Metrics.} We provide quantitative results on five evaluation metrics: PSNR, SSIM \cite{wang2004image}, mPSNR \cite{munkberg2006high}, MS-SSIM \cite{wang2003multiscale} and HDR-VDP-2.2.2 \cite{mantiuk2011hdr}. The PSNR and the mPSNR are error-based metrics. The SSIM and the MS-SSIM are structural metrics. Specifically, mPSNR is the multi-exposure peak signal-to-noise ratio, which is calculated by averaging the MSE values independently for different exposure levels. The HDR-VDP2 is used to measure performance degradation in luminance. Additionally, we adopt the inference speed and the number of parameters as metrics, which are denoted as Inf. Time and Params in Tabel 1. Note that the inference time for 2$\times$ and 4$\times$ upsampling is computed on a LR SDR of size 1080$\times$1920 and 540$\times$960 with an NVIDIA TITAN Xp GPU, respectively.

\vspace{1mm}
\noindent\textbf{Quantitative Comparison.} The qualitative results, speed and parameters are summarized in Table~\ref{tab1}. A more intuitive representation is provided in Figure \ref{fig1}. As shown in Table~\ref{tab1}, our proposed approach achieves state-of-the-art performance and speed. In particular, GPGMNet outperforms JSI-GAN, a divide and conquer methods, by up to 0.87 dB in PSNR for factor 4, while having about 70$\%$ fewer parameters and about 3 times faster speed. When compared with the previous state-of-the-art method JSInet, our proposed method still outperforms by 0.64dB in PSNR and preserves the lead in speed and number of parameters. Furthermore, compared with other methods, GPGMNet improves by 0.83 dB in mPSNR, 0.0043 in SSIM, 0.0032 in MS-SSIM, and 0.78 in HDR-VDP. In order to further verify the generalization and effectiveness of our method, we selected the test set of the HDRTV1K dataset, which contains 4K SDR frames and 4K HDR frames pairs. To make a paired test set for the Joint SR-ITM task, we made the test set by downsampling the 4K SDR frames on the test set of HDRTV1K dataset \cite{chen2021new} with bicubic filtering. Experimental results show that our method can achieve outstanding performance in PSNR and SSIM metrics on the test set of HDRTV1K dataset. Specifically, our proposed GPGMNet improves the PSNR by 0.72 dB and 0.70 dB over the previous state-of-the-art methods JSInet and JSI-GAN, respectively, as shown in Table~\ref{tab2}. The experimental results show that our method has certain generalization and remains effective on other dataset.

\noindent\textbf{Qualitative Comparison.} To be fair, we convert the 4K HDR images from the YUV-channel to the RGB-channel in PNG format encoded with 16-bits according to the ITU-R BT.2020-2 \cite{series2015parameter}, and then directly show them on the screen. Note that they "look hazy" compared to the corresponding images encoded by gamma EOTF, since the 4k HDR images are encoded by PQ EOTF and decoded by gamma EOTF on the SDR screen. The qualitative comparison of the proposed GPGMNet is given in Figure \ref{fig5}. Previous state-of-the-art methods tend to handle poorly in color mapping, structural consistency, and artifact removal. In particular, as shown in the first, second and third rows, our proposed GPGMNet can accurately complete the color mapping. The fourth and fifth rows show that our model preserves the structural similarity in spatial locations better than previous methods. The third and sixth rows show that GPGMNet can generate results closer to the ground truth without any artifacts. Overall, our approach has better visual results than the previous state-of-the-art methods.

\vspace{1mm}
\noindent\textbf{Comparison with the deep Cascade Method.} For comparison with the cascade method, we directly connect the state-of-the-art SR method, RLFN, and the state-of-the-art ITM method for SDRTV-to-HDRTV problem, HDRTVNet \cite{chen2021new}. To be fair, we only choose the first-stage network of HDRTVNet, named AGCM by chen et al. \cite{chen2021new}, since its the second-stage network can be seen as equivalent to EDSR and the third-stage networks of the HDRTVNet will cause a reduction in PSNR. We take LR-SDR images as input to the cascaded network, denoted as RLFN+AGCM and AGCM+RLFN. We held both networks at the same order of magnitude as our proposed GPGMNEt and training them on the same dataset using the same strategy in endd-to-end manner. We conduct quantitative comparisons on the 4K HDR dataset \cite{kim2019deep}, and the results are shown in Table~\ref{tab3}. Our proposed GPGMNet improves the PSNR by 0.51 dB and 0.50 dB over RLFN+AGCM and AGCM+RLFN, respectively. Experimental results show that our proposed GPGMNet is more efficient. 

\subsection{Color Comparison}


\begin{table}[t]
\centering
\caption{Mean $L_2$ error in L*a*b space, lower is better. Our proposed GPGMNet is capable of leraning  color and luminance mapping better than previous state-of-the-arts work. All results are calculated on 4K HDR dataset.}
\label{tab4}
\setlength{\tabcolsep}{6mm}{
\begin{tabular}{ccc}
\toprule
\multirow{2}{*}{Method} & \multicolumn{2}{c}{MSE$\downarrow$}             \\ \cmidrule{2-3} 
                        & \multicolumn{1}{c}{L*a*b}   & L-only \\ \midrule
JSInet                  & \multicolumn{1}{c}{20.31} & 6.99  \\ \midrule
JSI-GAN                 & \multicolumn{1}{c}{21.59} & 7.43  \\ \midrule
\textbf{GPGMNet(Ours)} & \multicolumn{1}{c}{\textbf{17.01}} & \textbf{5.32}  \\ 
\bottomrule
\end{tabular}}
\end{table}

\begin{figure}[t]
  \centering
  \includegraphics[width=\linewidth]{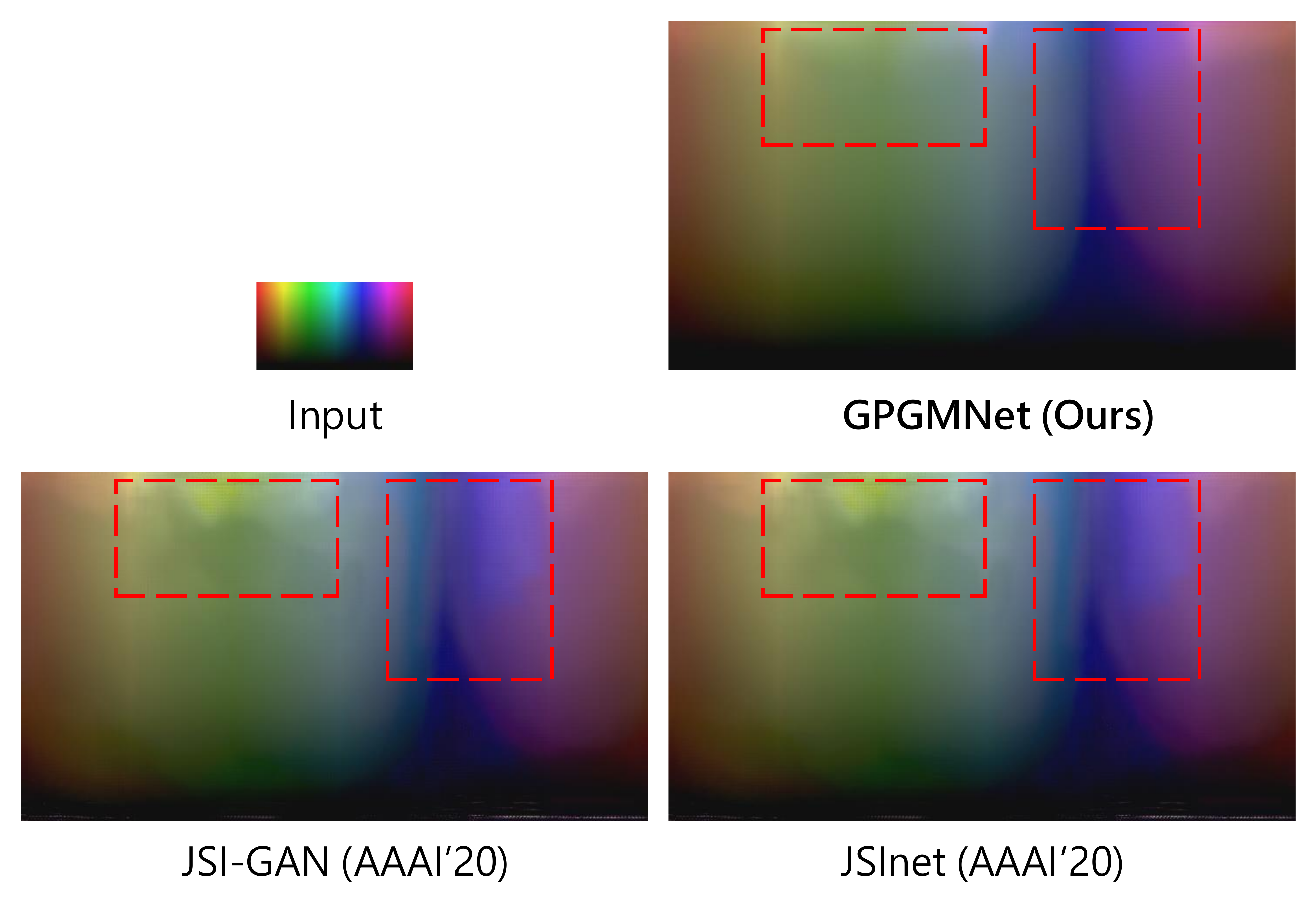}
  \vspace{-0.25in}
  \caption{Color Transition Test. We implemented a color transition test using a low-resolution standard dynamic range color bar map as input. Compared to previous state-of-the-art methods, our method produces more natural result with smooth transition.}
  \label{fig6}
\end{figure}

\begin{figure*}[h]
  \centering
  \includegraphics[width=\linewidth]{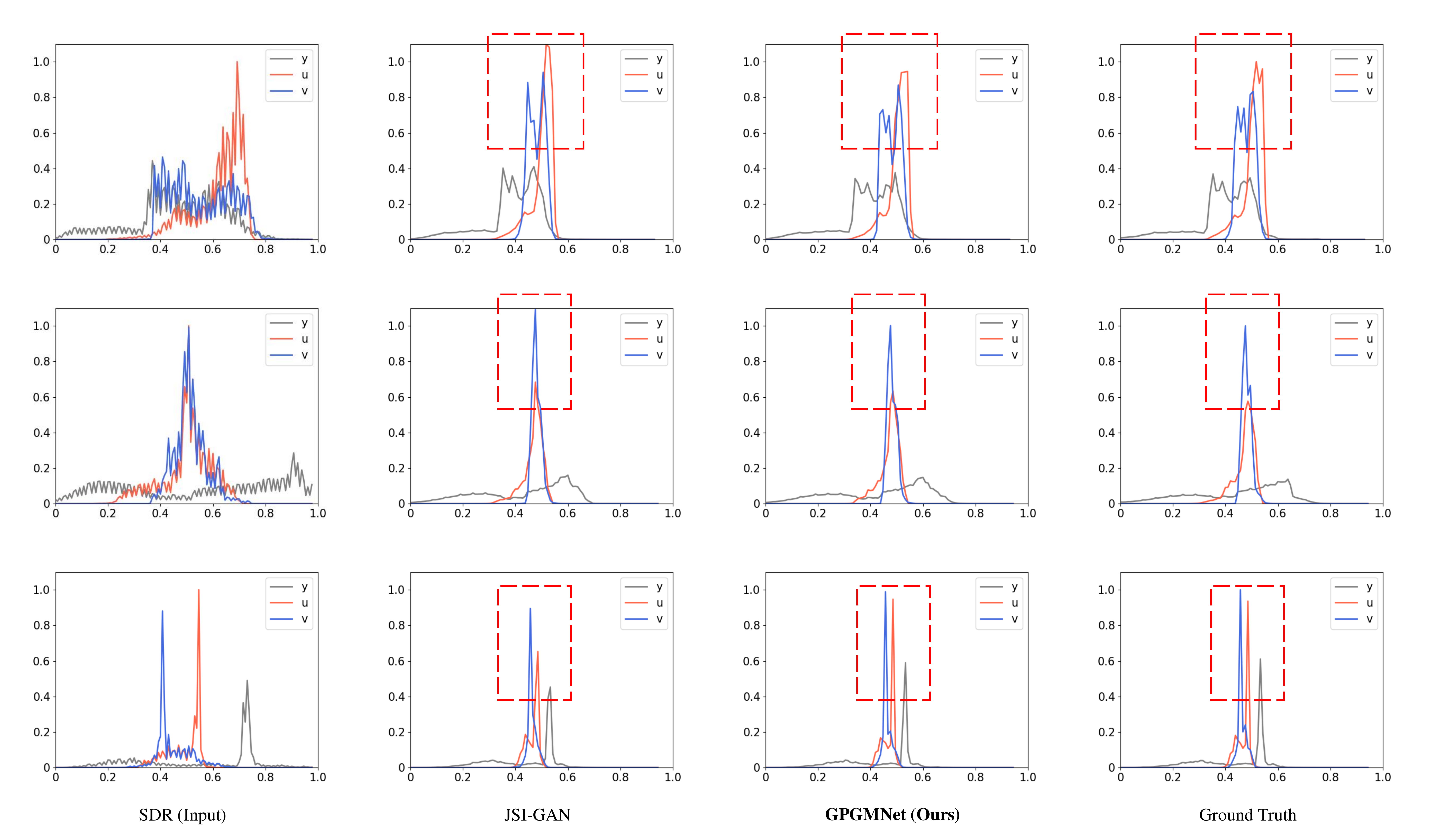}
  \vspace{-0.25in}
  \caption{Quantitative comparison of color histograms on YUV-channel. The horizontal axis represents the grayscale value (divided into 128) and the vertical axis represents the number of pixels (normalized by Ground Truth).}
  \vspace{-0.1in}
  \label{fig7}
\end{figure*}

We observe that previous methods perform poorly in maintaining holistic color conformity and smooth color transitions. To reveal this phenomenon, we conduct color testing also by visual qualitative comparison and quantitative comparison. First, we perform a color transition test using a man-made color bar as the input image following chen et al. \cite{chen2021new}. As shown in Figure \ref{fig6}, the outputs of the previous state-of-the-art methods JSInet and JSI-GAN are prone to unnatural artifacts and color blending problem. Second, we use color histograms to test the color distribution of the images generated by our method and JSI-GAN, respectively. As shown in Figure \ref{fig7}, the color distribution of the results generated by our method is closer to that of ground truth. In addition, due to the 8 bit depth limitation, it can be observed that SDR frames do not have a smooth transition in the horizontal axis compared to HDR frames. Based on the above observation, our method can achieve better recovery of color information and smooth transitions. For quantitative evaluation, we compute the mean in Lab space following \cite{gharbi2017deep}. The report of mean $L_2$ error in L*a*b color space (lower is better) is shown in Table~\ref{tab4}. In particular, our proposed method reduces the MSE by 3.3 and 1.67 over JSInet in Lab space and L-only, respectively. Experimental results suggest that our proposed method is capable of leraning color and luminance mapping better than previous state-of-the-arts work.

\subsection{Ablation Study}
We designed two sets of ablation studies for the global priors extraction module and the joint modulated residual module, respectively.

\begin{table}[t]
  \centering
  \caption{Ablation study on the global priors extraction module.}
  \label{tab5}
  \begin{threeparttable} 
        \begin{tabular}{lccccc}
            \toprule
            Variant &a &b &c &d &e\\
            \midrule
            CCP (w/o GF) &$\surd$ &$\times$ &$\times$ &$\surd$ &$\times$\\
            \midrule
            CCP (w/ GF) &$\times$ &$\surd$ &$\times$ &$\times$ &$\surd$\\
            \midrule
            SSP &$\times$ &$\times$ &$\surd$ &$\surd$ &$\surd$\\
            \midrule
            PSNR (dB) &34.01 &34.07 &34.18 &34.25 &\textbf{34.38}\\
            \bottomrule
        \end{tabular}
        
    \begin{tablenotes}
        \item[$*$]w/ GF denotes with guided filtering
        \item[$*$]w/o GF denotes without guided filtering
    
    \end{tablenotes}
  \end{threeparttable}
\end{table}

\vspace{1mm}
\noindent\textbf{Global Priors Extraction Module.} We performed a set of ablation experiments in GPGMNet by retraining different variants of the module. We trained the CCP and the SSP extraction branches to compare with the global priors extraction module, respectively. Additionally, we also verify the performance of using guided filtering in the CCP extraction branch. Table~\ref{tab5} shows the PSNR performance for different combinations of several variants. As shown in columns a, c, and d of Table~\ref{tab5} and columns b, c, and e of Tabel~\ref{tab5}, employing dual-branch bring gains in PSNR compared to single branch. The reason for the gain is that 
dual-branch extracts more global prior information, such as the color conformity prior and the structural similarity prior. As shown in columns d and e of Table~\ref{tab5}, the guided filtering brings a gain of 0.13 in the PSNR. This is because the guided filtering enables the CCP extraction branch to focus on extracting color conformity prior and low-frequency information in a favour of ITM, and keeps the SSP extraction branch more intensively to optimize structural similarity prior. Experiments suggest that the global priors can be extracted more fully when using CCP and SSP extraction branches simultaneously.

\begin{table}[t]
  \centering
  \caption{Ablation study on the GSMB.}
  \label{tab6}
  \begin{threeparttable} 
        \begin{tabular}{lcccc}
            \toprule
            Block &SFT &GFM$^1$ &GFM$^2$ &GSMB (ours)\\
            \midrule
            PSNR (dB) &34.06 &33.96 &34.17 &\textbf{34.38}\\
            \bottomrule
        \end{tabular}
        
    \begin{tablenotes}
        \item[$*$]$^1$ denotes the GFM layer withf condition network from AGCM.
        \item[$*$]$^2$ denotes the GFM layer with our proposed GPEM.
    \end{tablenotes}
    
  \end{threeparttable}
\end{table}

\begin{table}[t]
  \centering
  \caption{Ablation study on the SPCB.}
  \label{tab7}
  \begin{threeparttable} 
        \begin{tabular}{ccccc}
            \toprule
            Variant &GSMB &SA &SPCB &PSNR (dB) \\
            \midrule
            a &$\surd$ &$\surd$ &$\times$ &34.25\\
            \midrule
            b &$\surd$ &$\times$ &$\surd$ &\textbf{34.38}\\
            \bottomrule
        \end{tabular}
  \end{threeparttable}
\end{table}

\vspace{1mm}
\noindent\textbf{Joint Modulated Residual Module.} We set up a set of ablation experiments for our proposed joint modulated residual modules by retraining different feature modulation layers. For SFT, we reproduce it following Chen et al. \cite{chen2021hdrunet}. We compare with the GFM layers with GSMB in a consistent global condition network, denoted as GFM$^2$. To be fair, we also compare with the GFM layer with the condition network from AGCM, denoted as GFM$^1$. As shown in Table~\ref{tab6}, our proposed GSMB outperforms both SFT and GFM in PSNR, thanks to the preservation of spatial location information while performing global modulation. In addition, the comparison between GFM$^1$ and GFM$^2$ further demonstrates the advanced performance of our proposed GPEM. To verify the effectiveness of our proposed SPCB module, we compare the SPCB with the spatial attention (SA) module \cite{woo2018cbam}. As shown in Table~\ref{tab7}, the SPCB achieves superior performance compared to the SA.

    

\section{CONCLUSION}

In this paper, we have proposed GPGMNet, a CNN-based Joint SR-ITM framework, which consists of GPEM and JMRMs. The GPEM extracts global priors via SSP and CCP extraction branches for SR and ITM tasks, respectively. A GSMB is devised to simultaneously modulate the multi-scale spatial feature from the SPCB and global priors by scaling and shifting operations. The JMRMs jointly restores texture features and color information and modulate the feature maps by GSMBs. Qualitative and quantitative results show that our proposed method outperforms other state-of-the-art methods with lower computational cost. Despite the outstanding performance achieved by our method in 4K HDR dataset, there is currently no an actual benchmark dataset for the Joint SR-ITM task. Our future work will produce suitable datasets for the Joint SR-ITM task by inviting professional content producers to create HDR videos and their corresponding SDR videos from RAW format videos respectively. In addition, the exploitation of multi-frame information or time-domain correlation in mapping LR-SDR to 4K-HDR is one of our future works.

\bibliographystyle{IEEEtran}
\bibliography{IEEEabrv,main}

\vfill

\end{document}